# Learner to Learner Fuzzy Profiles Similarity Using a Hybrid Interaction Analysis Grid

Chabane Khentout[1*], Khadidja Harbouche[1], Mahieddine Djoudi[2]

[1] Computer Science Department and LRSD Laboratory, University Ferhat Abbes of Setif, Setif 19000, Algeria
[2] Computer Science Department and Techne Labs, University of Poitiers, 86073 Poitiers Cedex 9, France

Corresponding Author Email: chabane.khentout@univ-setif.dz

**ABSTRACT**

The analysis of remote discussions is not yet at the same level as the face-to-face ones. The present paper aspires two-fold. On the one hand, it attempts to establish a suitable environment of interaction and collaboration among learners by using the speech acts via a semi structured synchronous communication tool. On the other, it aims to define behavioral profiles and interpersonal skills hybrid grid by matching the BALES' IPA and PLETY's analysis system. By applying the fuzzy logic, we formalize human reasoning and, thus, giving very appreciable flexibility to the reasoning that use it, which makes it possible to take into account imprecisions and uncertainties. In addition, the educational data mining techniques are used to optimize the mapping of behaviors to learner's profile, with similarity-based clustering, using Eros and PCA measures. In order to show the validity of our system, we performed an experiment on real-world data. The results show, among others: (1) the usefulness of fuzzy logic to properly translate the profile text descriptions into a mathematical format, (2) an irregularity in the behavior of the learners, (3) the correlation between the profiles, (4) the superiority of Eros method to the PCA factor in precision.

## 1. INTRODUCTION

The learning systems are complex and, often unpredictable. They are designed by exploiting, combining and implementing multidisciplinary theories <<psychology, cognitive sciences, didactics, ergonomics, social sciences, IT … >>.

The COVID-19 pandemic has delivered an unprecedented shock to education systems, disrupting the lives of nearly 1.6 billion pupils and students in more than 190 countries on all continents. Lockdown has heavily affected the lives 94% of the world's educated population, and up to 99% in low- and lower-middle-income countries [1]. Shifting attention to remote learning has therefore become inevitable, necessitating urgent and radical changes in an aspiration to protect populations and curb the spread of the pandemic. Nevertheless, several challenges can hinder the practicality of distant learning. For instance, the content of the interactions during a working session, the participation and motivation levels of students as well as the exchanges structuring.

The inevitable global introduction of distance education during the covid-19 pandemic must encourage the establishment of inclusive education systems to overcome the shortcomings inherent in sole reliance on in-class education. More specifically, it ought to boost the potential of individuals and promote collective development in all areas of life, including teaching and learning. Algeria is no exception to this constatation.

In order to draw up a "Learning Profile in Algeria » a study encompassing 112 students and 24 teachers in the Department of Computer Science at the University Sétif (UFAS1, Algeria) was conducted [2]. This study should allow us, among other things, to situate ourselves in relation to existing learning models in order to propose an adequate material solution to the problems relating to the distant quality of learning such as the sociological isolation of the learner, the loss of motivation, etc. The survey results show that learners assimilate knowledge in different ways. Some prefer group work others assimilate better individually [2]. According to this study, a large majority of the learners have no difficulties collaborating (77%). Seven in ten learners (70%) prefer the group work to the individual one, and 67% assimilate better in a group. Among the scrutinized learners, 66% found that an interaction between learners is fruitful on the educational plan.

The partisans of group work find that this mode constitutes the basis of the collective intelligence. They argue that it allows them to accelerate the thinking and the understanding process, to learn to empathetically listen to others, to exchange information, to communicate effectively, to test and improve their capacities and ideas, to overcome their inadequacies and to share the knowledge and work methods. They find that interactions among learners is pedagogically fruitful. As a result, they openly agree to collaborate with other learners [2].

Through teamwork, collaborative learning is achieved more adequately. A team is perceived as a group of people interacting actively to carry out a common target. It necessitates the strategic distribution of tasks and the convergence of his members' efforts [3].

Lewin [4] defined the group cohesion as the willingness of individuals to stick together, and believed that without cohesiveness a group could not exist: the group is a whole which is not reduced to the sum of its parts. It establishes, with its closet circles, a dynamic structure (a field), the main elements of which are the subgroups, the members, the communication channels, the barriers … In this case, the mutual ignorance of the participants is not desirable.

In an attempt to directly or indirectly help students as participants in technologically mediated activities, the teachers as observers of these activities and the tutor as assistant and academic counselor, the computer-based interactions analysis is deployed (see Figure 1). The main contributions of this work are: (1) Drawing up a profile of the learners using a hybrid analysis grid by combining Bale's IPA (Interaction Process Analysis) and PLETY grid, (2) fuzzy analysis of the interactions (3) calculating the similarity between behavioral profiles evolving over time of pairs of learners using PCA factor and Eros method, (4) grouping together learners with similar behaviors using hard and soft clustering, (5) experimenting the system on real-world data.

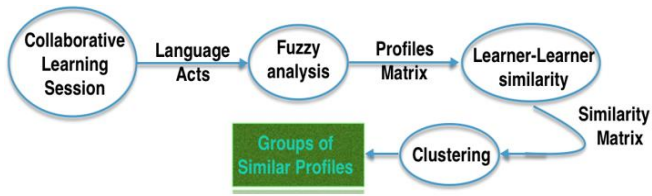

**Figure 1.** Synoptic of the interaction analysis process

The remainder of the paper is organized as follows: initially, we present the related works in the area of the learners interactions analysis, in section2. In section 3, we present our hybrid analysis grid combining both Bales IPA and PLETY system. Section 4 details the proposed approach. The experimental of our system are discussed in Section 5. Section 6 concludes the paper.

## 2. RELATED WORKS

In a socio-constructivist approach, interactions between learners play a dynamic role in individual learning [5]. Nonetheless, the analysis of remote discussions is not yet at the same level as the face-to-face ones. Evidently, the retrieved data tends to be more imperfect than those obtained from direct interaction. In this line, several studies have focused on the implementation of automatic systems for analyzing traces of interactions.

The Bales IPA has been used in some research in the area of Computer Sciences. Birnholtz et al. [6], for example, presented an experimental study of two-people groups (dyads) editing documents together. The focus of the study, was to analyze group maintenance, impression management and relationship-focused behavior. Savolainen [7] studied the extent to which blogs are used as interactive forums in which people can ask and share information by considering four reactions of the IPA categorization: showing positive and negative reactions, and asking and answering questions.

There are some related works that rely on IPA classifications but these ones are made manually by human experts. Kim et al. [8], for example, manually classified video recordings segments into IPA categories. Löfstrand and Zakrisson [9] followed a similar approach with the aim of identifying competitive and non-competitive behavior.

Approaches such as Academic Talk [10], Group Leader Tutor [11] and EPSILON [12] attempted to reduce the effort needed to encode group dynamics into IPA categories by introducing the concept of *sentence openers*. Sentence openers predefine a limited set of sentences that are allowed to start a communication act. Each sentence opener is directly matched to one IPA category.

Contrary to Bales, a limited number of researchers have relied on PLETY's research on ethology. George [13] has proposed an environment called SPLASH that uses the PLETY grid to analyze mediated synchronous conversations.

The present review of the literature allows us to highlight the differences with our approach that will be presented in this article:
1. We use a hybrid grid (Bales and PLETY) which allows us to automatically classify learners according to behavioral and relational profiles.
2. Instead of using sentence openers and limiting the IPA categories, we'll use the intentions of the exchanged messages, not their contents, through positive and negative speech acts.
3. To improve the accuracy of the classification, we use a heuristic method to calculate the coefficients of the profiles.
4. The use fuzzy logic for the analysis of that is closet to human reasoning.
5. Introduction of data mining methods for the measurement of similarity between learners and grouping learners with the same profiles.

## 3. LEARNERS INTERACTIONS ANALYSIS GRIDS

### 3.1 BALES IPA and PLETY grid

The learners interactions analysis is the process of automatic or semi-automatic analysis, based on data obtained by the participants' own activity, with the aim of understanding the activity mediated by technology. This understanding allows human or even artificial stakeholders to participate in the control of the activity, helping them with awareness, self-assessment, or self-regulation.

Many psychosociologists were interested in the study of the interactions in a workgroup.

- The study led by Robert F. Bales, from 1946 to 1949, within the framework of the group dynamics [14], allowed him, after observation of several small newsgroups face to face, to finalize a system of Interaction Process Analysis (IPA) [15].

This IPA supplies a tool which allows on one hand to make a quantitative and qualitative analysis of the interpersonal relations and the dynamics of the group and on the other hand to establish individual profiles, to analyze the quality of the socio-emotional relations and the behavior directed to the work from the number and of kind of interactions. This tool is called: the grid of Bales. The principle of application of the IPA method in on-line groups is very close to that deployed for the face to face groups. It is a classification of behavioral acts (act by act), to analyze the data in order to obtain descriptive indications of the functioning of the group and extract factors influencing this process.

The grid of Bales is a set of 12 categories which allow to describe, according to their nature, positive aspects of the intervention, and conversely negative aspects [15].

- Robert PLETY, researcher at the communications ethology laboratory of the university Lumière - Lyon 2

(France), studied a lot the behavior of pupils working in groups; he analyzed in particular interactions between pupils working in a group of four on the resolution of algebra problems [16]. To achieve that purpose, he started from a micro-analysis of the sparring and gestural exchanges in the group to end in the determination of behavior profiles of the pupils. From this analysis, PLETY defined behavior profiles which characterize the roles played by the partners in the group.

From these observations, PLETY highlights four (04) typical profiles of behavior: the organizer, the verifier, the seeker and the independent [13].

According to this study, these four profiles find themselves in almost all the analyzed groups (16 in all). PLETY brings an element of answer, speaking about learners groups distributed on network, by underlining that he met, curiously, the same aspects of membership, cohesion and leadership there that in the ordinary groups [13].

### 3.2 Hybrid interaction analysis grid

The question that arises, is to know how to transpose the results of Bales and PLETY works into a conversation context, using IT (Information Technologies)?

The grid which we propose arises from the crossing of both Bales IPA and PLETY analysis system, by considering only the following situations:
- Distance Learning.
- Ignoring any socio-emotional relation between learners.
- Ignoring the communicative gestures.
- Drawing up a behavioral and relational learner's profile.

Table1 summarizes our hybrid grid arises after having rearrange both grids of Bales and PLETY.

The hybrid grid gives us a picture of the social behavior of a learner during a collaborative learning session.

**Table 1.** Hybrid grid

|  | Interventions | | Driven reactions | | PV |
|---|---|---|---|---|---|
|  | Acts+ | Acts- | Acts+ | Acts- |  |
| **Organizer** | P | E | A, M | - | Large |
| **Verifier** | A, M | D, E | - | S, C | Large |
| **Seeker** | - | E | M |  | Small |
| **Independent** | - | S |  | S, C | Low |
| **Collaborator** | Orientation aspect | | Decision aspect | | |
|  | Acts+ | Acts- | Acts+ | Acts- |  |
|  | P, M | E, C | A | D, S |  |

P: To propose  A: To approve  D: To disapprove C: To decline
E: To elucidate S: To stand mute M: To demonstrate
Acts+: Positive acts Acts-: Negative acts PV: Participation volume

It underlines two categories of profiles:
- **Behavioral profiles:** Including 5 PLETY patterns, which are determined by:
  - The type of intervention
  - The driven reactions to an intervention
  - The intervention volume.

These patterns are:

- *The organizer* who intervenes by his proposals and his questions a lot, the other members react generally positively to its interventions.
- *The verifier* who reacts to the various proposals and answers the questions of his peers. Other members react little to his interventions.
- *The seeker:* He does not intervene too much but he always tries to understand by asking questions which are accepted well by his peers.
- *The independent*: His proposals or evaluations are rare, even non-existent. His peers do not react to his interventions which remain pending.

- **Relationship profile:** Inspired by Bales IPA, we have here the collaborative profile of the learner by means of the *orientation* and the *decision* aspects of the learner. These two aspects are described by positive and negative acts.

## 4. PROPOSED FUZZY PROFILES SIMILARITY USING HYBRID INTERACTION ANALYSIS GRID

### 4.1 Behavior fuzzy analysis

The fuzzy logic is an extension of Boolean logic, introduced with the 1965 proposal of « fuzzy set theory » by Lotfi Zadeh. It gives very appreciable flexibility to the reasoning that use it, which makes it possible to take into account imprecisions and uncertainties [17].

One of the interests of fuzzy logic to formalize human reasoning, is that the rules are expressed and stated in natural language. This is appropriate with our desire to draw up a learner profile in the group according to the discussions made during a collaboration session, although it is always difficult to try to model human behavior.

The fuzzy system [18] is divided into Five steps: (1) the input, (2) the Fuzzification, (3) the Inference engine, (4) the Defuzzification (5) the Output.

For each of the behavioral profile pattern (Organizer, Verifier, Seeker and Independent) and for the relational profile (Collaboration) we will do a fuzzy analysis what will generate 5 fuzzy systems (as shown in Figure 2).

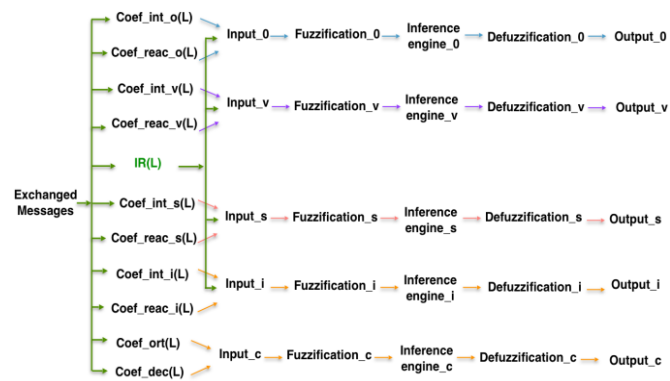

**Figure 2.** Synoptic view of the fuzzy systems

### Step1: Profiles coefficients

To analyze automatically the conversations between learners, we have to calculate some heuristic formulas using the indications given above and the messages exchanged between learners. In reality we'll use the intentions of the exchanged messages and not their contents, by means of positive and negative speech acts [19].

➢ For the relationship profile, and for each learner, we'll calculate the orientation and the decision indices given by the formulas: Eq. (1) and Eq. (2).

The Orientation Index (*Ind_ort_p (L)*):

$$Ind\_ort(L) = \frac{\sum Act\_pos\_ort(L)}{\sum Act\_neg\_ort(L)} \quad (1)$$

where:
- *Act_pos_ort* is the number of orientation positive acts,
- *Act_neg_ort* is the number of orientation negative acts.

The Decision Index (*Ind_dec_p (L)*):

$$Ind\_dec(L) = \frac{\sum Act\_pos\_dec(L)}{\sum Act\_neg\_dec(L)} \quad (2)$$

where:
- *Act_pos_dec* is the number of decision positive acts,
- *Act_neg_dec* is the number of decision negative acts.

➢ From the description of each behavioral profile p (where $p \in \{o, v, s, i\}$ for respectively organizer, verifier, seeker, independent) we'll calculate the intervention and reaction coefficients and the intervention ratio given by the formulas: Eq. (3), Eq. (4) and Eq. (5).

The Intervention Coefficient (*Coef_int_p(L)*): We see that the intervention coefficient of each profile is a ratio between the number of participations that characterize this profile and the total number of participations of the learner during a session.

$$Coef\_int\_p(L) = \frac{\sum Act\_int\_p(L)}{\sum Act\_int(L)} \quad (3)$$

where:
- *Act_int_p* is the number of positive and negative intervention acts according to the profile p,
- *Act_int* is the total number of intervention acts.

The Reaction Coefficient (*Coef_reac_p(L)*): We see that the reaction coefficient of each learner is a ratio between the reactions of his peers to the interventions during a session.

$$Coef\_reac\_p(L) = \frac{\sum Act\_reac\_p(L)}{\sum Act\_int\_p(L)} \quad (4)$$

where:
- *Act_reac_p* is the number of positive and negative reaction acts according to the profile p,
- *Act_int_p* is the number of positive and negative intervention acts according to the profile p.

The Intervention Ratio (*IR(L)*): It is the learner's participation or interventions (*Act_int(L)*) rate compared to the group's average participation during a session (*Avr_int_grp*).

$$IR(L) = \frac{\sum Act\_int(L)}{\sum Avr\_int\_grp} \quad (5)$$

**Step 2: The fuzzification**
The second step involves construction of membership functions. It corresponds to the identification of the linguistic variables. Two profiles' analysis are to be implemented:

**Step 2- Stage 1:** Behavioral PLETY analysis:
We have three inputs for each of the profiles "Organizer, Verifier, Seeker and Independent" with as variable, the interaction coefficient for input1, the reaction coefficient for input2 and the intervention or participation ratio for input 3.

In light of the profiles' characteristics shown in the hybrid grid the following fuzzy sets (linguistic variables) are proposed:

- The organizer input1 = {Passive, Active}
- The organizer input2 = {Negative, Positive}
- The verifier input1 = {Indifferent, Interested}
- The verifier input2 = {Little, Enough}
- The seeker input1 = {Incurious, Curious}
- The seeker input2 = {rejected, Accepted}
- The independent input1 = {Present, Absent}
- The independent input2 = {Heard, Disregarded}
- The behavioral profiles input3= {Low, Average, Important}

The membership function of input1 as shown in Figure 3, for all the behavioral profiles, is a sigmoid function defined on the interval [0 .. 1] and given by the formula Eq. (6):

$$f1(x) = 1/(1 + e^{\pm 14(x-0.5)}) \quad (6)$$

The membership function f2 of input 2 (Eq. (7)), for the behavioral profiles «Organizer, Independent and Seeker», is a sigmoid function too (Figure3), defined on the interval [0 .. 1]:

$$f2(x) = 1/(1 + e^{\pm 14(x-0.5)}) \quad (7)$$

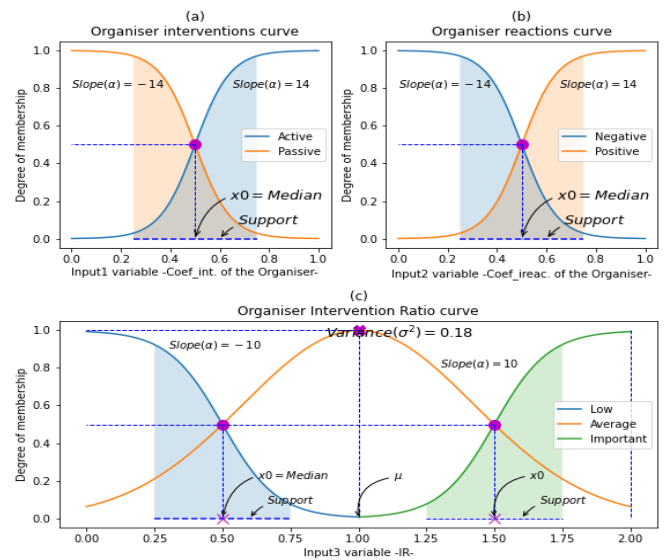

**Figure 3.** Linguistic variables of the organizer

The membership function $f_v2$ of input 2, for the Verifier profile is a gaussian function in the interval [0..0.5] and a sigmoid function in the interval [0.25..1] (Eq.(8)) as shown in Figure 4:

$$f_v2(x) = \begin{cases} 1/e^{-(x-0.25)/0.02} \\ 1/(1+e^{\pm 11(x-0.625)}) \end{cases} \quad (8)$$

The membership function f3 of input3 (Figure3 and Figure4), for every behavioral profile is defined by Eq. (9) as follow:

$$f3(x) = \begin{cases} 1/e^{-(x-1)/0.2} \\ 1/(1+e^{-10(x-1.5)}) \\ 1/(1+e^{10(x-0.5)}) \end{cases} \quad (9)$$

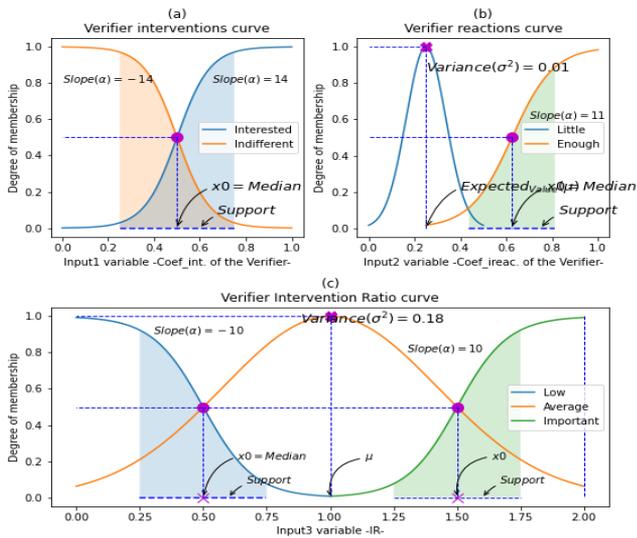

**Figure 4.** Linguistic variables of the verifier

For the Output, the following fuzzy sets are proposed:
- The organizer, the verifier and the seeker output = {Weak_p, Insufficient_p, Medium_p, Satisfactory, Good_p} where p ∈ {o, v, s}
- The independent output = {Integrated, Upper_accepted, Accepted, Less-accepted, Isolated}

The membership function (Figure 5), for all the behavioral profiles is defined as a gaussian function (Eq. (10)) on the interval [0 .. 12]:

$$F(x) = 1/e^{-(x-\mu)^2/0.8} \quad (10)$$

where, μ is the expected value.

**Step 2- Stage 2:** Relational analysis
It concerns the Collaborator profile. The latter has three variables: orientation index for input1, decision index for input 2 and collaboration index for output.
The fuzzy sets proposed for this profile are:
- The collaborator input1 = {Weak, Good}
- The collaborator input2 = {Weak, Good}
- The collaborator output= {Weak_c, Average_c, Good_c /c = collaborator}.

The Input1 and input2 membership functions (Figure 6) are both sigmoid defined on the interval [0 .. 2] (Eq. (11)):

$$fc1(x) = fc2(x) = 1/(1+e^{\pm 7(x-1)}) \quad (11)$$

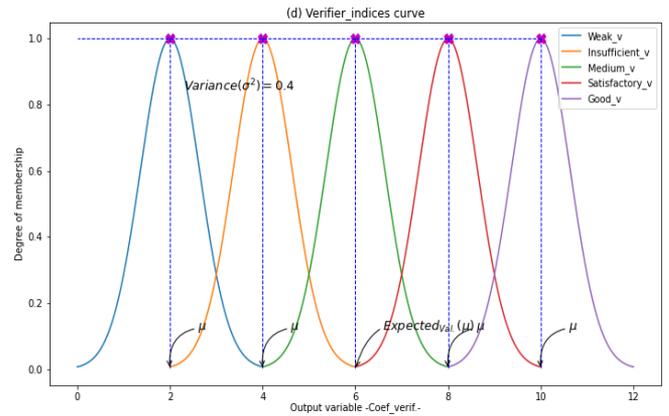

**Figure 5.** Output Linguistic variables of the verifier

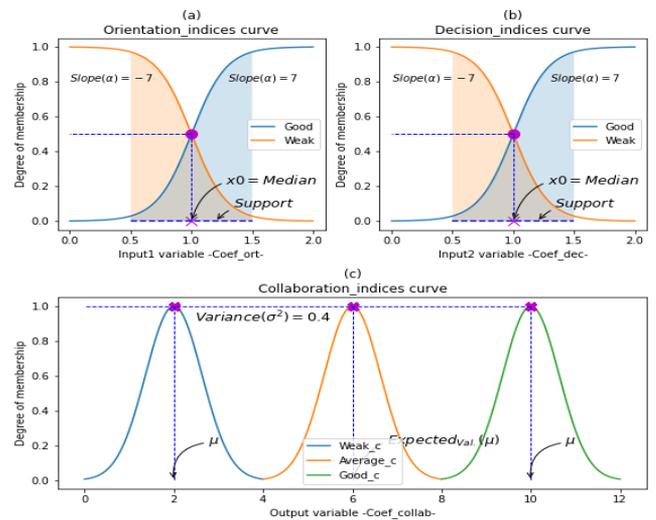

**Figure 6.** Linguistic variables (Orientation, Decision and Collaboration)

For the Output, the membership function as shown in Figure 6, is a gaussian function (Eq. (12)) defined in the universe of discourse X = [0..12]:

$$Fc(x) = 1/e^{-(x-\mu)^2/0.8} \quad (12)$$

**Step 3: The inference engine**
The third step involves developing a set of rules for the knowledge base.
Given below are the set of rules using IF-THEN logic.
The f**uzzy knowledge base** is made up of all the fuzzy rules according to each profile. Let's take as an example the two profiles: the organizer and the collaborator.

**Step 3- Stage 1: Definition of fuzzy rules**
The organizer inference rules:

1. **If** *Coef_int_o* is Active and *Coef_reac_o* is Positive and *IR* is Important **then** *Coef_organization* is Good_o
2. **If** *Coef_int_o* is Active and *Coef_reac_o* is Positive

and *IR* is Average **then** *Coef_organization* is Satisfactory_o
3. **If** *Coef_int_o* is Active and *Coef_reac_o* is Negative and *IR* is Important **then** *Coef_organization* is Satisfactory_o
4. **If** *Coef_int_o* is Passive and *Coef_reac_o* is Positive and *IR* is Important **then** *Coef_organization* is Satisfactory_o
5. **If** *Coef_int_o* is Active and *Coef_reac_o* is Positive and *IR* is Low **then** *Coef_organization* is Medium_o
6. **If** *Coef_int_o* is Passive and *Coef_reac_o* is Positive and *IR* is Average **then** *Coef_organization* is Medium_o
7. **If** *Coef_int_o* is Active and *Coef_reac_o* is Negative and *IR* is Average **then** *Coef_organization* is Medium_o
8. **If** *Coef_int_o* is Active and *Coef_reac_o* is Negative and *IR* is Low **then** *Coef_organization* is Insufficient_o
9. **If** Coef_int_o is Passive and Coef_reac_o is Positive and IR is Low **then** Coef_organization is Insufficient_o
10. **If** Coef_int_o is Passive and Coef_reac_o is Negative and IR is Important **then** Coef_organization is Insufficient_o
11. **If** Coef_int_o is Passive and Coef_reac_o is Negative and IR is Average **then** Coef_organization is insufficient_o
12. **If** Coef_int_o is Passive and Coef_reac_o is Negative and IR is Low **then** Coef_organization is Weak_o The collaborator inference rules:
13. **If** Ind_ort is Good and Ind_dec is Good **then** Ind_collab is Good_c
14. **If** Ind_ort is Good and Ind_dec is Weak **then** Ind_collab is Average_c
15. **If** Ind_ort is Weak and Ind_dec is Good **then** Ind_collab is Average_c
16. **If** Ind_ort is Weak and Ind_dec is Weak **then** Ind_collab is Weak_c

**Step 3- Stage 2: Selection of fuzzy operators**

The choice of the fuzzy operators allows to determine **the inference engine,** which is generated in our case by using the Min/Max Zadeh operators [12].

Let S be the set of ordered pairs (fuzzy variables, crisp variables).

$S$ = {(*Coef_int_o*(Active), *Xo1*), (*Coef_int_o*(Passive), *Xo2*), (*Coef_reac_o*(Positive), *X'o1*), (*Coef_reac_o*(Negative), *X'o2*), (*IR*(Important), *I1*), (*IR*(Average), *I2*), (*IR*(Low), *I3*), (*Coef_organization*(Good_o), *Yo1*), *Coef_organization* (Satisfactory_o), *Yo2*), (*Coef_organization*(Medium_o), *Yo3*), (*Coef_organization*(Insufficient_o), *Yo4*), (*Coef_organization* (Weak_o), *Yo5*), (*Ind_orient*(Good), *X1*), (*Ind_orient*(Weak), *X2*), (*Ind_dec*(Good), *X'1*), (*Ind_dec*(Weak), *X'2*), (*Ind_collab*(Good_c), *Y1*), (*Ind_collab*(Average_c), *Y2*), (*Ind_collab*(Weak_c), *Y3*)}.

**Step 3- Stage 3: Apply fuzzy rules**

We apply the fuzzy rules, quoted above, with the MIN/MAX operators and obtain:
The organizer:
*Yo1* = Min (*Xo1, X'o1, I1*)
*Yo2* = Max (Min (*Xo1, X'o1, I2*), Min (*Xo1, X'o2, I1*), Min (*Xo2, X'o1, I1*))
*Yo3* = Max (Min (*Xo1, X'o1, I3*), Min (*Xo2, X'o1, I2*), Min (*Xo1, X'o2, I2*))
*Yo4* = Max (Min (*Xo1, X'o2, I3*), Min (*Xo2, X'o2, I1*), Min (*Xo2, X'o2, I2*), Min (*Xo2, X'o1, I3*))
*Yo5* = Min (*Xo2, X'o2, I3*)
The collaborator:
*Y1* = Min (*X2, X'2*)
*Y2* = Max (Min (*X1, X'1*), Min (*X2, X'2*))
*Y3* = Min (*X1, X'1*)

**Step 4: The defuzzification**

The Mean of Maxima (MOM) method is applied here [20]. In this method, the defuzzied value is taken as the element with the highest membership values. When there are more than one element having maximum membership values, the mean value of the maxima is taken.

Let p a behavioral profile, p ∈ {o, v, s, i}.
*Yp* = Max (*Yp1, Yp2, Yp3, Yp4, Yp5*).
*Y* = Max(*Y1, Y2, Y3*).

We calculate the inputs (or abscissas) named $A_p$, $A$ corresponding to the outputs $Y_p$, $y$ respectively:

$$Y = e^{(A-\mu)^2/0.8} \qquad (13)$$

$$Y_p = e^{(A_p-\mu_p)^2/0.8} \qquad (14)$$

where, $\mu \in \{2, 6, 10\}$ and $\mu p \in \{2, 4, 6, 8, 10\}$

Let $A$ be a fuzzy set with membership function $Y(x)$ defined over $x \in X$, where X is the universe of discourse.

The defuzzied value is let say $x^*$ of a fuzzy set is defined by the formula Eq. (15):

$$x^* = (\sum_{x_i \in M} x_i) / |M| \qquad (15)$$

where:
- M = { xi| Y ( xi ) is equal to the height of the fuzzy set A},
- |M| is the cardinality of the set M.

**Step 5: The output decision**
It corresponds to the definitive decision:
1. Giving the relational profile of the learner according to Bales with the percentages corresponding to the values Good_c, Average_c and Weak_c and the final collaboration index A, corresponding to the mean of MOM method.
2. Giving the behavioral profile of the learner according to PLETY with the percentages corresponding to the values Weak_p, Insufficient_p, Medium_p, Satisfactory_p and Good_p for each of the profiles Organizer, Verifier and Seeker, as well as the final indices: Animation, Check and Quest; calculated by the mean of MOM method.
3. Giving the behavioral profile of the learner according to PLETY with the percentages corresponding to the values Integrated, Upper_accepted, Accepted, Less-accepted and Isolated for the Independent, as well as the final index Independence; calculated by the mean of MOM method.

**4.2 Learner-to-learner behavior similarity**

The analysis of interactions between learners is based on data obtained by the participants' own activity during collaborative sessions. The learner profiles calculated for each of these sessions allow us to obtain a multivariate time series (MTSLP = Multivariate Time Series for Learner Profiles). Each MTS [21] concerns a learner. It is a series of final indices $x_i(t)$; [$i = 1, .., n; t = 1, \ldots, m$], calculated sequentially through discrete time (collaborative sessions) where i indexes the calculation made at each time point t, thereby n = 5, since the columns of the matrix represent the collaboration, animation, check, quest and independence indices, calculated above (Figure 7).

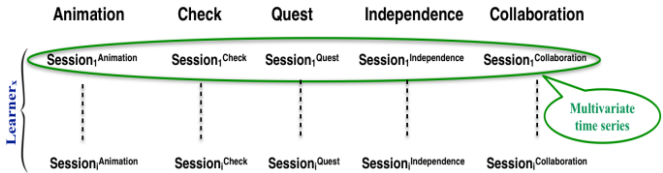

**Figure 7.** Synoptic of an MTSLP

An MTSLP:
- should be treated as a whole, since there are usually important correlations among the variables in MTS data,
- may not be transformed into one long univariate time series.

**Step 6: Learner to learner similarity**

In this section, we aim to calculate the learner-to-learner similarity which provides a way of quantifying the degree of agreement between two learners behaviors.

The multidimensional similarity measure aims to indicate the level of similarity between several databases or groups of data simultaneously.

To measure the similarity between two learners $L_A$ and $L_B$, we consider their MTSLP A, B of respective dimension $d_A \times 5$ and $d_B \times 5$.

These matrices (MTSLPs) have the same number of columns (5), but not necessarily the same number of rows, because:
- a learner within a group can be absent during a collaborative session,
- the number of work sessions may differ from one group to another.

We will use two methods dealing with Multivariate Time Series MTS:
- Principle Component Analysis Similarity factor (PCAS).
- The Eros method (Extended Frobenius norm).

Both methods apply to principal components of matrices. Hence, we need to pre-process our data. Algorithm1 computes the principal component of two MTSLPs:

1. Apply the SVD (Singular Value Decomposition) on the $M_A$ and $M_B$ covariance matrices, and we obtain the decomposition of each into (U Σ VT). V is called the right eigenvector matrix, and U the left eigenvector matrix.
2. Consider $V_A$ and $V_B$, the two right eigenvector matrices resulting from the SVD. $V_A$ and $V_B$ are expressed in the following form: $V_A = [a1, \cdots, an]$ and $V_B = [b1, \cdots, bn]$, with ai and bi column orthonormal vectors of size n=5.

Because SVD is applied to covariance matrices, the eigenvectors and the principal components are used interchangeably [22].

**Algorithm 1** Computing the principal component of a multivariate time series of a learner A profiles $MTSLP_A$
**Input:** An n x 5 $MTSLP_A$, where n is the number of sessions and 5 is the number of profiles
**Output:** The right eigenvector matrix $V_A$ and the left eigenvector matrix $U_A$
1: MA ← Covariance ($MTSLP_A$)
2: UA Σ VA ← SVD (MA)

**Step 6.1: Learner to learner similarity using PCA factor**

PCAS computes the similarity between the first k principal components [23].

1. PCAS firstly obtains the k principal components for each matrix A and B.
2. Intuitively, PCAS measures the similarity between two matrices by computing the squared cosine values between all the combinations of the n principal components from two matrices using Eq. (16).

$$PCAS(A, B) = \sum_{i=1}^{k} \sum_{j=1}^{k} \cos^2 \theta_{ij} \qquad (16)$$

where: $\theta_{ij}$ is the angle between the $i^{th}$ principal component of A and the $j^{th}$ principal component of B.

3. The range of PCAS is between 0 and n, to obtain a range between 0 and 1 we divide PCAS by n.

The result of the calculations carried out during this step between each pair of learners, will make it possible to obtain the learner-to-learner similarity matrix X as shown in Figure 8.

**Step 6.2: Learner to learner similarity using EROS method**

The intuition behind this method lies in its ability to process a different set of observations for each group of data with the same number n of variables (5 profiles).

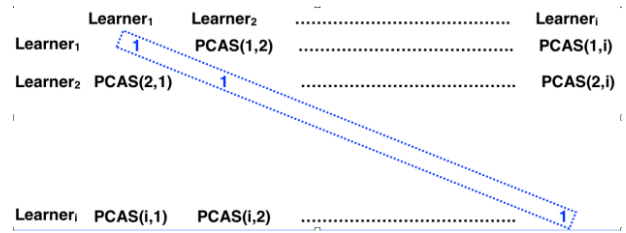

**Figure 8.** Learner to Learner PCA similarity matrix X

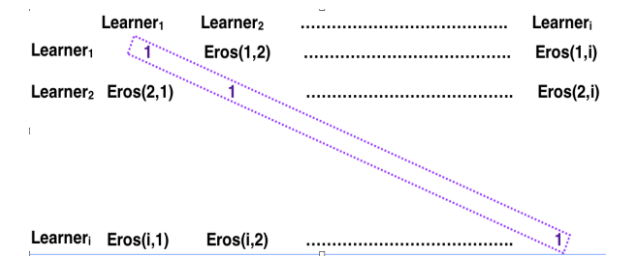

**Figure 9.** Learner to Learner Eros similarity matrix Y

This method is freed from the dimension problem by considering not the group of data but the eigenvalues and eigenvectors of the covariance matrix which, themselves, are of identical sizes [24].

Another advantage provided by this method concerns the dimension reduction of the data.

The Eros similarity between two MTSLP A and B; to obtain the learner-to-learner similarity matrix Y as shown in Figure 9; is given by:

$$Eros(A, B, w) = \sum_{i=1}^{n} w_i |\cos \theta_i| \quad (17)$$

where:
- $\cos \theta_i$ is the angle between $a_i$ and $b_i$,
- w is a weight vector with:

$$\sum_{i=1}^{n} w_i = 1 \text{ and } w_i \geq 0 \quad (18)$$

w vector is obtained by:
1. Taking the eigenvalues obtained from all the MTSLP items in the dataset.
2. Aggregating the eigenvalues obtained into one vector using the mean function.
3. Normalizing the weights ($\sum w_i = 1$ for $i \in [1..n]$) by applying the Eq. (19).

$$w_i \leftarrow w_i / \sum_{j=1}^{n} w_j \quad (19)$$

The range of Eros is between 0 and 1, with 1 being the most similar.

### 4.3 Learners' profiles clustering

We'll use two clustering approaches: soft clustering and hard clustering, and we'll focus on three main methods:

**Step 7.1: Hierarchical Ascendant Classification (HAC)**

It is a technique whereby an indexed hierarchy is constructed among the elements of the data set [25]. Elements are successively "merged," that is, subsumed into entities or clusters comprising two or more elements, based on their distance in factor space, and on application of a merging rule. As distance, the generalized Euclidean distance in factor space is normally used (Eq. (20)). In accordance with the merging rule, each merger yields an index of similarity, which can be used to construct a tree that depicts the hierarchical relationships. Such a tree is called a dendrogram [26].

$$d(x_i, y_i) = \sqrt{\sum_{i=1}^{n} (x_i - y_i)^2} \quad (20)$$

**Step 7.2: K-means**

Unlike HAC methods, which are difficult to implement on large datasets, K-means [27] is a clustering algorithm (algorithm2) which is widely used in such situations. It is an iterative algorithm that proceeds as follows:

**Algorithm 2 Computing the K-means clusters**
**Input: An n x 5 Learners profiles matrix, where n is the number of learners and 5 is the number of profiles.**
**Output: K learners clusters**
1. Fix the number of clusters k
2. Selection of k clusters to generate
3. Initialization of centroids with random values
4. Repeat
5. Calculating the distance between the objects and the center of gravity of the cluster using Eq. (20)
6. Assignment of points to the nearest centroid.
7. Displacement of the centroid to the cluster mean.
8. Until there is no change in the center of the clusters.

The Euclidean distance, given by Eq. (20), is generally considered to determine the distance between each data object and the centers of the cluster.

The Euclidean distance (d) between two vectors X = {$x_1$, $x_2$,..., $x_n$} and Y = {$y_1$, $y_2$,..., $y_n$} is given by:

When all data objects are included in some clusters, the first step is complete and early grouping is performed. This iterative process continues until the criterion function is minimized.

**Step 7.3: Fuzzy c-means**

The application of fuzzy logic for various scientific and technical goals has been commented on for decades [28]. This approach differs from the classical hard clustering where each object of the data set finds its own cluster. Thus, an object either belongs to a defined cluster or is out of it. The application of Fuzzy theory to the problem of finding similarity between objects of interest leads to the conclusion that a particular object can belong simultaneously to more than one cluster, but with different degrees of membership (DOMs) between 0 and 1 [29].

To calculate the weights and determine the clusters of a set of data points X = {x1, x2, x3 ..., xn} we perform the steps of Algorithm 3. Firstly, we randomly select a set of centroids V = {v1, v2, v3 ..., vc}, then, for each data point and for each cluster, we calculate the fuzzy membership of a data i to the cluster j using Equ. 21. Next, we calculate the new fuzzy centers $v_j$ using Equ. 22. We repeat the calculation of the fuzzy memberships and the fuzzy new centers until the change in the center of the clusters between two iteration steps k and k+1 is less than the termination criterion β ($\|Z^{(k+1)} - Z^{(k)}\| < β$) or I iterations value is achieved.

$$z_{ij} = \frac{1}{\sum_{k=1}^{c} \left(\frac{\|x_i - v_j\|}{\|x_i - v_k\|}\right)^{2/(m-1)}} \quad (21)$$

$$v_j = \frac{\sum_{i=1}^{n} z_{ij}^m \cdot x_i}{\sum_{i=1}^{n} z_{ij}^m} \quad (22)$$

where:
- m is the fuzziness index $m \in [1..\infty]$,
- c represents the number of cluster center,
- $z_{ij}$ represents the membership of $i^{th}$ data to $j^{th}$ cluster center,
- n is the number of data points,
- $v_{ij}$ represents the $j^{th}$ cluster center and c represents the number of cluster center.

| Algorithm 3 Computing the C-means clusters |
|---|
| **Input:** An n x 5 Learners profiles matrix, where n is the number of learners and 5 is the number of profiles |
| **Output:** C learners clusters with fuzzy membership |
| 1.     Fix the number of clusters c |
| 2.     Fix the termination criterion β |
| 3.     Fix the number of maximum iterations I |
| 4.     Fix the fuzziest index m |
| 5.     Let X = {x1, x2, x3 ..., xn} be the set of data points |
| 6.     Let V = {v1, v2, v3 ..., vc} be the set of centers |
| 1.     Randomly select (c) cluster centers |
| 2.     k ← 1 |
| 3.     repeat |
| 4.     for j=1 to c do |
| 5.      for i=1 to n do |
| 6.     Calculate the fuzzy membership $z_{ij}$ using Equ. 21 |
| 7.     end for |
| 8.     end for |
| 9.     Compute the new fuzzy centers $v_j$ using Equ. 22 |
| 10.    k ← k + 1 |
| 11.    until $\|W^{(k+1)} - W^{(k)}\| < \beta$ or k=I+1 |

## 5. EXPERIMENTS AND RESULTS

As we were faced with a lack of adequate datasets to evaluate and test our system, we collected our own data from 28 second year undergraduate students, from the computer science department of the science faculty, at Ferhat Abbas University of Setif, over a period of two months. These students were divided into groups of 4 learners, meeting remotely for collaborative work: this is a mini-project on RISC processors. The interactions of the learners in terms of the number of language acts were collected. After the data pre-processing, the experiment process is as following: we will calculate the different coefficients of the profiles using fuzzy logic, then we will proceed to the calculation of the similarity between pairs of learners using the two methods EROS and PCA. The next step is to group learners together using the three clustering methods: HAC, FCM, and K-means. After each step we will discuss the results obtained.

### 5.1 Collaborative session structuring

We recommend a centralized structure within the group. The **coordinator** "team leader" is the only member having the right to make a decision after consulting the other members.

We use the participation model to structure our collaborative learning. During a working session the team leader exposes the problem then invites every present member to give its opinion. The team leader intervenes once the round table discussion is ended. He takes notes and presents the clear conclusions: if all the members approve the proposal, the problem is considered as solved, if instead the members disapprove of the proposal then a new proposal to be discussed is made.

The members can ask to the initiator of the solution to clear up it better. The initiator can decline the request of clarification as he can explain or demonstrate his solution what returns us to a new evaluation phase.

Time constraints make that in case of not consensus, the team leader adopts a proposal as a solution.

The finite-state machine of Figure 10 allows to illustrate the various states of the collaborative session where every interaction imposes the aforesaid constraints to answer a previous interaction or activate other interactions.

- Any exchange structured into speech acts between the members of a group during a collaborative session is detected by the profiler. The purpose of the profiler is to identify the nature of each intervention and not its content.
- It assigns a score which signifies the number of uses of an act per learner, and collects the acts performed by the members of the group.
- The acts collected by type are saved in a dataset.

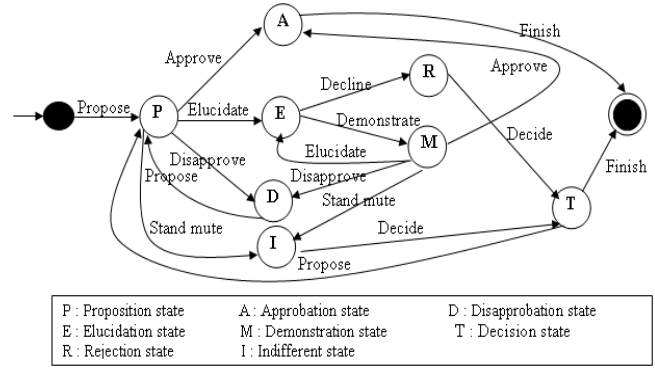

P : Proposition state    A : Approbation state    D : Disapprobation state
E : Elucidation state    M : Demonstration state    T : Decision state
R : Rejection state    I : Indifferent state

**Figure 10.** Illustration by a finite state machine of the collaborative session states

### 5.2 Results and discussion

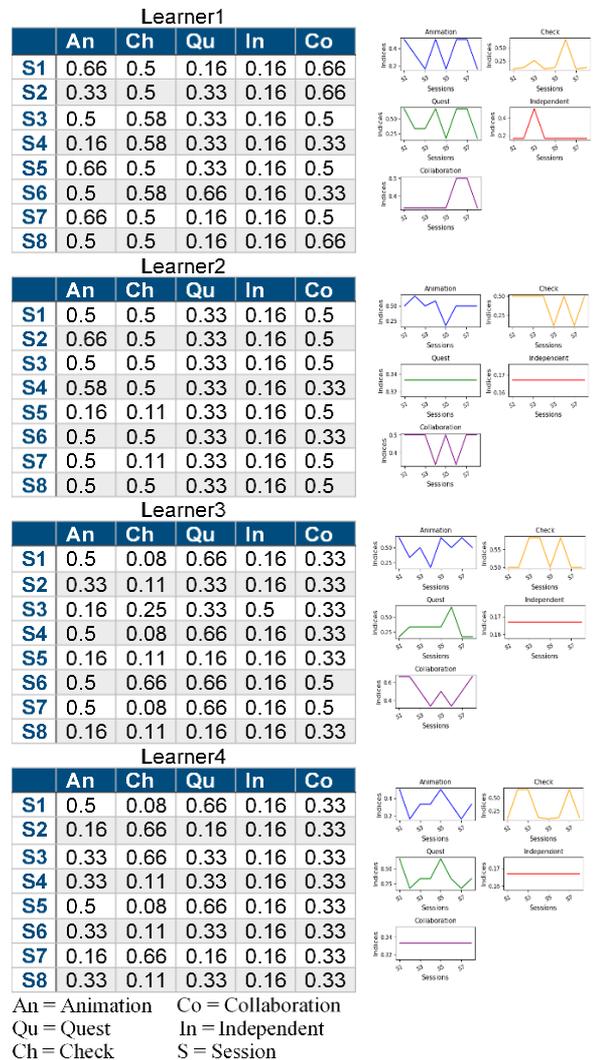

Learner1

| | An | Ch | Qu | In | Co |
|---|---|---|---|---|---|
| S1 | 0.66 | 0.5 | 0.16 | 0.16 | 0.66 |
| S2 | 0.33 | 0.5 | 0.33 | 0.16 | 0.66 |
| S3 | 0.5 | 0.58 | 0.33 | 0.16 | 0.5 |
| S4 | 0.16 | 0.58 | 0.33 | 0.16 | 0.33 |
| S5 | 0.66 | 0.5 | 0.33 | 0.16 | 0.5 |
| S6 | 0.5 | 0.58 | 0.66 | 0.16 | 0.33 |
| S7 | 0.66 | 0.5 | 0.16 | 0.16 | 0.5 |
| S8 | 0.5 | 0.5 | 0.16 | 0.16 | 0.66 |

Learner2

| | An | Ch | Qu | In | Co |
|---|---|---|---|---|---|
| S1 | 0.5 | 0.5 | 0.33 | 0.16 | 0.5 |
| S2 | 0.66 | 0.5 | 0.33 | 0.16 | 0.5 |
| S3 | 0.5 | 0.5 | 0.33 | 0.16 | 0.5 |
| S4 | 0.58 | 0.5 | 0.33 | 0.16 | 0.33 |
| S5 | 0.16 | 0.11 | 0.33 | 0.16 | 0.5 |
| S6 | 0.5 | 0.5 | 0.33 | 0.16 | 0.33 |
| S7 | 0.5 | 0.11 | 0.33 | 0.16 | 0.5 |
| S8 | 0.5 | 0.5 | 0.33 | 0.16 | 0.5 |

Learner3

| | An | Ch | Qu | In | Co |
|---|---|---|---|---|---|
| S1 | 0.5 | 0.08 | 0.66 | 0.16 | 0.33 |
| S2 | 0.33 | 0.11 | 0.33 | 0.16 | 0.33 |
| S3 | 0.16 | 0.25 | 0.33 | 0.5 | 0.33 |
| S4 | 0.5 | 0.08 | 0.66 | 0.16 | 0.33 |
| S5 | 0.16 | 0.11 | 0.16 | 0.16 | 0.33 |
| S6 | 0.5 | 0.66 | 0.66 | 0.16 | 0.5 |
| S7 | 0.5 | 0.08 | 0.66 | 0.16 | 0.5 |
| S8 | 0.16 | 0.11 | 0.16 | 0.16 | 0.33 |

Learner4

| | An | Ch | Qu | In | Co |
|---|---|---|---|---|---|
| S1 | 0.5 | 0.08 | 0.66 | 0.16 | 0.33 |
| S2 | 0.16 | 0.66 | 0.16 | 0.16 | 0.33 |
| S3 | 0.33 | 0.66 | 0.33 | 0.16 | 0.33 |
| S4 | 0.33 | 0.11 | 0.33 | 0.16 | 0.33 |
| S5 | 0.5 | 0.08 | 0.66 | 0.16 | 0.33 |
| S6 | 0.33 | 0.11 | 0.33 | 0.16 | 0.33 |
| S7 | 0.16 | 0.66 | 0.16 | 0.16 | 0.33 |
| S8 | 0.33 | 0.11 | 0.33 | 0.16 | 0.33 |

An = Animation    Co = Collaboration
Qu = Quest    In = Independent
Ch = Check    S = Session

**Figure 11.** Evolution of four Learners' profiles

Emotions, the change in temperament and cognitive and sociological evolutions make analyzing human behavior a fairly challenging task. The study of educational systems, in particular the learners' behavior, is complex hindering the understanding the results. The deployment of data mining reveals more or less precise models. The study carried out enabled us to identify the results postulated infra:

1. As shown in Figure 11, we notice that:
- the behavior of the learners is not quite regular (it varies over time), that said, there is always a dominant profile more than the others and present in all the sessions;
- the groups are not balanced (each learner has a particular character).

2. Figure 12 shows some correlations between profiles:
- we can see, a weak positive relationship associates the animation profile to the check and quest ones. This is to be expected if we review the characteristics of each profile, e.g., the organizer intervenes by his proposals and his questions a lot, the other members react generally positively to its interventions, the verifier reacts to the various proposals and answers the questions of his peers and the seeker always tries to understand by asking questions which are accepted well by his peers.
- a curved cubic relationship linking the check profile to the collaboration and quest profiles;
- no relationship between the independent and the other profiles;
- we can see an outlier, due to an atypical character often present in groups (learner too strong or learner too weak).

These correlations are due to the interlacing and interconnection of the profiles characteristics.

3. Figure 13 shows that Eros and PCA give two learner-to-learner similarity indices. Recall that, in general, there are three types of transformation that should be considered for similarity measures, i.e., shift, scale and time warping. The similarity measures for time series should then be invariant to those transformations. Eros measures the similarity between two MTS items by comparing how far the principal components are apart using the aggregated eigenvalues as weights taking into account the variance for each principal component.

Thus, Eros gives the best precision and elapsed computing time than PCA [24].

4. We can see, that the groups are not really independent and that some learners can belong to several groups at the same time but with different degrees, as shown in Table 2.

5. The experimental results show a slightly difference between the hard clustering methods and the c-means one as shown in Figure 14. This result is expected and if the size of the dataset was larger, we would have found more differences. Because, one of the weaknesses of the chosen clustering methods is that they give varying results on different executions of an algorithm. A random choice of cluster models produces different results, which leads to inconsistency.

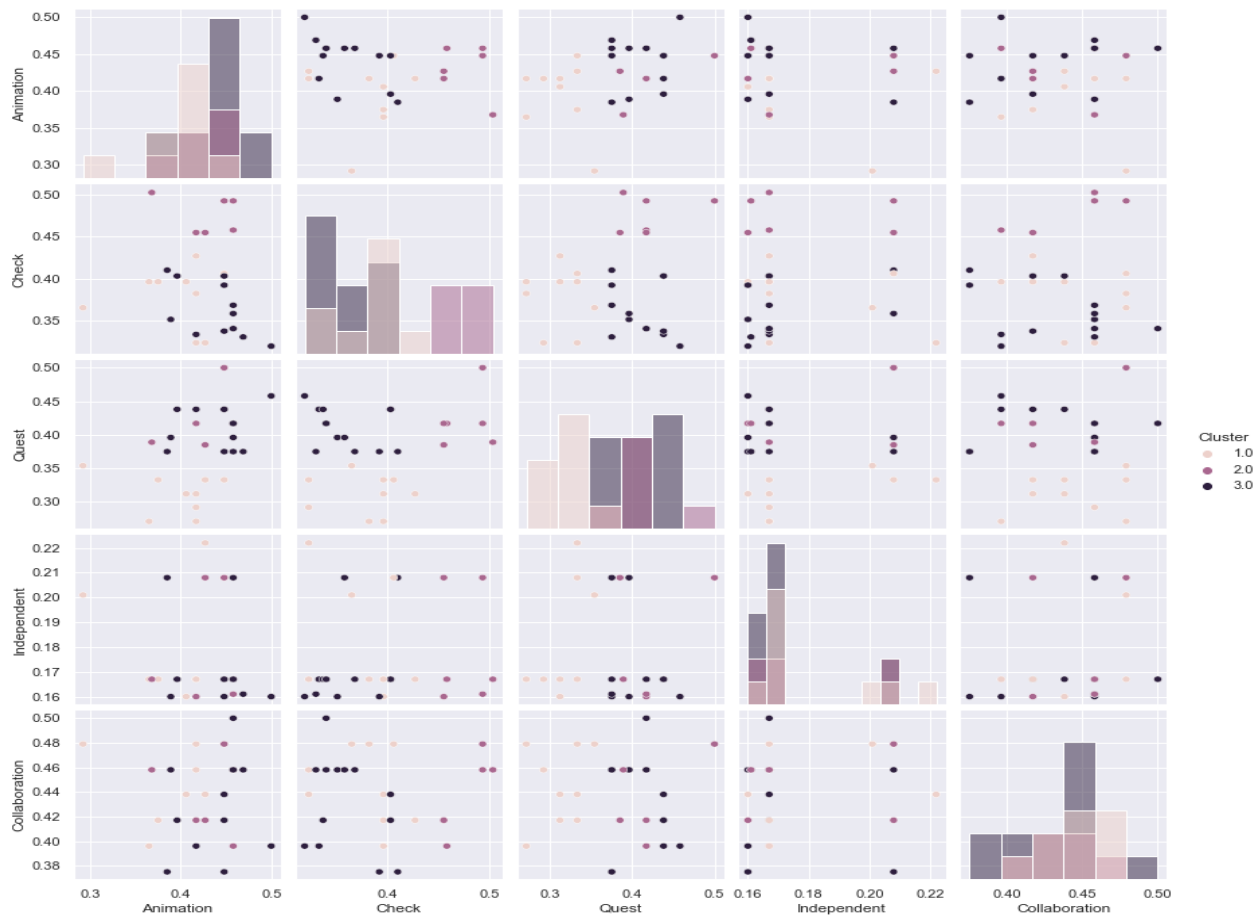

**Figure 12.** Profiles correlations

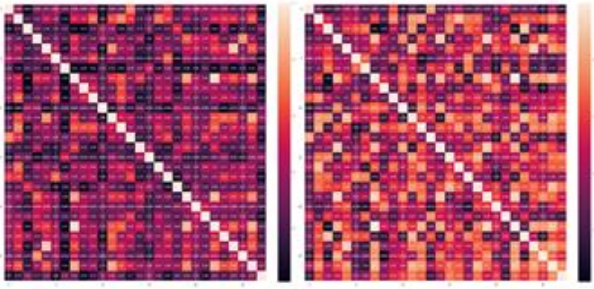

| (a) Eros similarity | (b) PCA similarity |

**Figure 13.** Learner to learner similarity

**Table 2.** Cluster-Learner membership

| Learner | FCM: Cluster's membership | | | HAC clusters | K-means clusters |
|---|---|---|---|---|---|
| | Cluster1 | Cluster2 | Cluster3 | | |
| 1 | 0,331 | 0,330 | **0,339** | 3 | 3 |
| 2 | 0,334 | **0,335** | 0,330 | 3 | 3 |
| 3 | 0,334 | 0,330 | **0,336** | 3 | 3 |
| 4 | 0,331 | **0,335** | 0,333 | 1 | 1 |
| 5 | 0,334 | 0,329 | **0,337** | 3 | 3 |
| 6 | 0,331 | **0,339** | 0,330 | 1 | 1 |
| 7 | **0,334** | 0,331 | 0,333 | 3 | 3 |
| 8 | 0,332 | 0,332 | **0,336** | 3 | 3 |
| 9 | **0,335** | 0,332 | 0,332 | 2 | 2 |
| 10 | 0,332 | **0,336** | 0,332 | 1 | 1 |
| 11 | **0,335** | 0,335 | 0,330 | 2 | 2 |
| 12 | 0,332 | **0,339** | 0,329 | 1 | 1 |
| 13 | 0,332 | **0,336** | 0,332 | 1 | 1 |
| 14 | **0,336** | 0,330 | 0,334 | 3 | 3 |
| 15 | 0,332 | 0,329 | **0,338** | 3 | 3 |
| 16 | 0,333 | **0,338** | 0,329 | 1 | 1 |
| 17 | 0,333 | **0,335** | 0,332 | 1 | 1 |
| 18 | 0,332 | 0,331 | **0,337** | 3 | 3 |
| 19 | 0,336 | 0,327 | **0,337** | 3 | 3 |
| 20 | **0,336** | 0,333 | 0,331 | 2 | 2 |
| 21 | **0,335** | 0,333 | 0,332 | 2 | 2 |
| 22 | 0,333 | 0,328 | **0,339** | 3 | 3 |
| 23 | 0,334 | 0,331 | **0,336** | 3 | 3 |
| 24 | 0,333 | 0,330 | **0,337** | 3 | 3 |
| 25 | **0,334** | 0,332 | 0,333 | 2 | 2 |
| 26 | 0,332 | **0,336** | 0,331 | 1 | 1 |
| 27 | 0,331 | **0,334** | 0,334 | 1 | 1 |
| 28 | 0,334 | **0,335** | 0,331 | 2 | 2 |

The distances used in clustering in most of the times do not actually represent the spatial distances. In general, the only solution to the problem of finding global minimum is exhaustive choice of starting points.

In our case, both K-means and HAC clustering produce fairly higher accuracy but the K-means method requires less computation. FCM clustering produces close results to K-means clustering, yet it requires more computation time because of the fuzzy measures calculations involved in the algorithm.

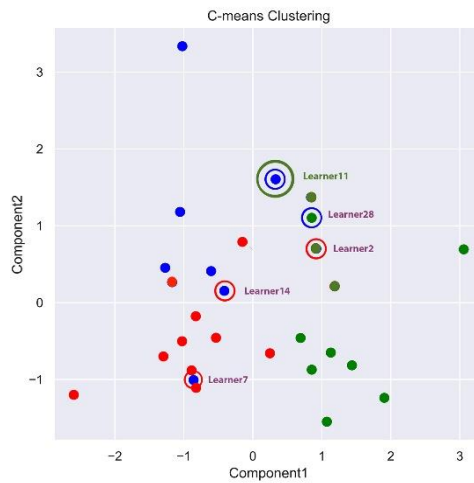

(a) Fuzzy C-means

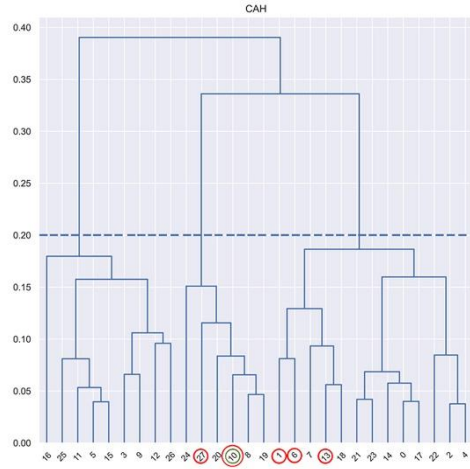

(b) Hierarchical clustering

**Figure 14.** Hard and soft clustering

## 6. CONCLUSION

The current study aimed to inspect collaborative learning. More specifically, it examined difficulties associated to collective work such as displaying a feeling of membership in the group, positive evolution of the learner and interaction among individuals. It proposed a tool which allows to provide a learner with a profile in a collaborative learning situation. The profiler uses heuristics indices given by the crossing of Bales grid and that of PLETY. The fuzzy logic supplied a mathematical formalism to implement these indices. The analysis should therefore allow the creation of groups of learners with similar behaviors using both Eros and PCA methods to calculate the similarity between two MTSLP.

The study would have been more comprehensive if larger data had been used. As a perspective, we would like to conduct the experiment on a larger population of learners and over a longer period while including several knowledge components. By means of other analysis grids and others computing ways, we aim improving the implemented tool. To deepen this topic, we propose the use of a deep learning model (CNN for example) to straighten and correct the classes predicted by the used methods, namely: FCM, K-means and HAC.

## REFERENCES

[1] United Nations. (2020). Policy brief: Education during COVID-19 and beyond.

https://www.un.org/development/desa/dspd/wp-content/uploads/sites/22/2020/08/sg_policy_brief_covid-19_and_education_august_2020.pdf

[2] Harbouche, K. (2013). Intelligent agent-based environment work for collaborative learning on the web. PHD thesis, Computer Science Department, Ferhat Abbas University.

[3] Khentout, C. (2006). Interfaces et assistance à l'apprenant en enseignement à distance. PHD thesis, Computer Science Department, Ferhat Abbas University.

[4] Lewin, K. (1992). La dynamique des groupes. Sciences humaines, 14: 10-11.

[5] Doise, W., Mugny, G., James, A.S., Emler, N., Mackie, D. (2013). The social development of the intellect. https://doi.org/10.1016/0191-8869(87)90192-9

[6] Birnholtz, J., Steinhardt, S., Pavese, A. (2013). Write here, write now! An experimental study of group maintenance in collaborative writing. In Proceedings of the SIGCHI Conference on Human Factors in Computing Systems, pp. 961-970. https://doi.org/10.1145/2470654.2466123

[7] Savolainen, R. (2011). Asking and sharing information in the blogosphere: The case of slimming blogs. Library & Information Science Research, 33(1): 73-79. https://doi.org/10.1016/j.lisr.2010.04.004

[8] Kim, M.S., Kim, Y.S., Kim, T.H. (2007). Analysis of team interaction and team creativity of student design teams based on personal creativity modes. In International Design Engineering Technical Conferences and Computers and Information in Engineering Conference, 48043: 55-67. https://doi.org/10.1115/DETC2007-35378

[9] Löfstrand, P., Zakrisson, I. (2014). Competitive versus non-competitive goals in group decision-making. Small Group Research, 45(4): 451-464. https://doi.org/10.1177/1046496414532954

[10] Mcalister, S., Ravenscroft, A., Scanlon, E. (2004). Combining interaction and context design to support collaborative argumentation using a tool for synchronous CMC. Journal of Computer Assisted Learning, 20(3): 194-204. https://doi.org/10.1111/j.1365-2729.2004.00086.x

[11] Israel, J., Aiken, R. (2007). Supporting collaborative learning with an intelligent web-based system. International Journal of Artificial Intelligence in Education, 17(1): 3-40.

[12] Goodman, B.A., Linton, F.N., Gaimari, R.D., Hitzeman, J.M., Ross, H.J., Zarrella, G. (2005). Using dialogue features to predict trouble during collaborative learning. User Modeling and User-Adapted Interaction, 15(1): 85-134. https://doi.org/10.1007/s11257-004-5269-x

[13] George, S. (2001). Apprentissage collectif à distance: splash: un environnement informatique support d'une pédagogie de projet (Doctoral dissertation, Le Mans).

[14] Fournier, M., Austin John, L. (2011). In: L'abécédaire des sciences humaines. chapter1. http://www.scienceshumaines.com/austin-john-l_fr_12618.html

[15] Fahy, P.J. (2006). Online and face-to-face group interaction processes compared using Bales' interaction process analysis (IPA). European Journal of open, distance and e-learning.

[16] PLETY, R. (1996). L'apprentissage coopérant, Ethologie et psychologie des communications. ARCI Presse Universitaire, Lyon, France.

[17] Godjevac, J. (1999). Idées nettes sur la logique floue. PPUR presses polytechniques et universitaires Romandes. Lausanne, Switzerland.

[18] Gacôgne, L. (2001). Logique floue et applications. In: Institut d'Informatique d'Entreprise. Evry, France.

[19] Searle, J.R. (1990). Consciousness, explanatory inversion, and cognitive science. Behavioral and Brain Sciences, 13(4): 585-596. https://doi.org/10.1017/S0140525X00080304

[20] Nodelman, U., Allen, C., Perry, J. (1995). Stanford encyclopedia of philosophy. Bryant University. 2006-07-23. Retrieved 2008-09-30. https://plato.stanford.edu/entries/logic-fuzzy

[21] Hoppner, F. (2002). Learning dependencies in multivariate time series. In Proc. of the ECAI'02 Workshop.

[22] Jackson, J.E. (2005). A user's guide to principal components. John Wiley & Sons. https://doi.org/10.2307/3172717

[23] Abdi, H., Williams, L.J. (2010). Principal component analysis. Wiley interdisciplinary reviews: computational statistics, 2(4): 433-459. https://doi.org/10.1002/wics.101

[24] Yang, K., Shahabi, C. (2004). A PCA-based similarity measure for multivariate time series. In Proceedings of the 2nd ACM international workshop on Multimedia databases, pp. 65-74. https://doi.org/10.1145/1032604.1032616

[25] Sasirekha, K., Baby, P. (2013). Agglomerative hierarchical clustering algorithm-a. International Journal of Scientific and Research Publications, 83: 83. https://doi.org/10.1007/978-1-4419-9863-7_100033

[26] Zhou, S., Xu, Z., Liu, F. (2016). Method for determining the optimal number of clusters based on agglomerative hierarchical clustering. IEEE transactions on neural networks and learning systems, 28(12): 3007-3017. https://doi.org/10.1109/TNNLS.2016.2608001

[27] Fahim, A.M., Salem, A.M., Torkey, F.A., Ramadan, M. (2006). An efficient enhanced k-means clustering algorithm. Journal of Zhejiang University-Science A, 7(10): 1626-1633. https://doi.org/10.1631/jzus.2006.A1626

[28] Taherpour, A., Cheshmeh Sefidi, A., Bemani, A., Hamule, T. (2018). Application of Fuzzy c-means algorithm for the estimation of Asphaltene precipitation. Petroleum Science and Technology, 36(3): 239-243. https://doi.org/10.1080/10916466.2017.1416632

[29] Zhang, Y., Wang, W., Zhang, X., Li, Y. (2008). A cluster validity index for fuzzy clustering. Information Sciences, 178(4): 1205-1218. https://doi.org/10.1016/j.ins.2007.10.004